\documentclass{article}

\usepackage{PRIMEarxiv}

\usepackage[numbers]{natbib}

\usepackage[utf8]{inputenc} 
\usepackage[T1]{fontenc}    
\usepackage{hyperref}       
\usepackage{url}            
\usepackage{booktabs}       
\usepackage{amsfonts}       
\usepackage{nicefrac}       
\usepackage{microtype}      
\usepackage{lipsum}
\usepackage{fancyhdr}       
\usepackage{graphicx}       
\usepackage[table]{xcolor}  
\usepackage{multirow}
\usepackage{amsmath}
\usepackage{bm}
\usepackage{makecell}
\usepackage{cleveref}
\usepackage{subcaption}
\usepackage{float}
\usepackage{placeins}
\graphicspath{{media/}}     

\usepackage{import}
\usepackage[acronym]{glossaries}
\usepackage{xcolor}
\usepackage{amsmath}
\usepackage{amsfonts}
\usepackage{hyperref}
\usepackage{tabularx}
\usepackage{graphicx}
\usepackage{booktabs}
\usepackage{caption}
\usepackage{subcaption}

\usepackage[nottoc]{tocbibind}
\usepackage{bm}
\usepackage{tikz}
\usepackage{xspace}
\usepackage{xcolor} 
\usepackage{url} 

\usepackage{placeins}

\newcommand{\add}[1]{\textcolor{black}{#1}}

\usepackage{soul}
\setstcolor{red}

\usepackage{algorithm} 
\usepackage{algpseudocode}

\newcommand{\PreserveBackslash}[1]{\let\temp=\\#1\let\\=\temp}
\newcolumntype{C}[1]{>{\PreserveBackslash\centering}p{#1}}
\newcolumntype{R}[1]{>{\PreserveBackslash\raggedleft}p{#1}}
\newcolumntype{L}[1]{>{\PreserveBackslash\raggedright}p{#1}}

\usepackage{color}

\newcommand{\manu}[1]{{\color{black}#1}}

\title{Domain Adaptation for Image Classification of Defects in Semiconductor Manufacturing} 

\author{
  Adrian Poniatowski \\
  Infineon Technologies AG, University of Bologna \\
  \texttt{adrianponiatowski@gmail.com} \\
  \And
  Natalie Gentner \\
  Infineon Technologies AG, University of Padova \\
  \texttt{natalie.gentner@infineon.com} \\
  \And
  Manuel Barusco \\
  University of Padova \\
  \texttt{manuel.barusco@phd.unipd.it} \\
  \And
  Davide Dalle Pezze \\
  University of Padova \\
  \texttt{davide.dallepezze@unipd.it} \\
  \And
  Samuele Salti \\
  University of Bologna \\
  \texttt{samuele.salti@unibo.it} \\
  \And
  Gian Antonio Susto \\
  University of Padova \\
  \texttt{gianantonio.susto@unipd.it} \\
}

\begin{document}
\maketitle

\begin{abstract}
In the semiconductor sector, due to high demand but also strong and increasing competition, time to market and quality are key factors in securing significant market share in various application areas. 
Thanks to the success of deep learning methods in recent years in the computer vision domain, Industry 4.0 and 5.0 applications, such as defect classification, have achieved remarkable success.
In particular, Domain Adaptation (DA) has proven highly effective since it focuses on using the knowledge learned on a (source) domain to adapt and perform effectively on a different but related (target) domain.
By improving robustness and scalability, DA minimizes the need for extensive manual re-labeling or re-training of models.
This not only reduces computational and resource costs but also allows human experts to focus on high-value tasks.
Therefore, we tested the efficacy of DA techniques in semi-supervised and unsupervised settings within the context of the semiconductor field.
Moreover, we propose the DBACS approach, a CycleGAN-inspired model enhanced with additional loss terms to improve performance.
All the approaches are studied and validated on real-world Electron Microscope images considering the unsupervised and semi-supervised settings, proving the usefulness of our method in advancing DA techniques for the semiconductor field.
\end{abstract}


\section{Introduction}\label{sec:intro}

\newacronym{ML}{ML}{Machine Learning}
\newacronym{SVM}{SVM}{Support Vector Machines}
\newacronym{CNN}{CNN}{Convolutional Neural Networks}
\newacronym{IC}{IC}{Integrated Circuits}
\newacronym{SEM}{SEM}{Scanning Electron Microscopes}

Semiconductors are constructed by linking circuit structures on various layers built upon a thin slice of semiconductor material called a wafer. 
 The creation of those layers, including their complex individual structures and patterns, is done in a sequential manner through a combination of the steps.
 The repetition of process steps and re-entry into the production flow is part of the advanced and highly specialized fabrication process.

\begin{figure*}[!h]
\centering
\includegraphics[width=0.8\textwidth]{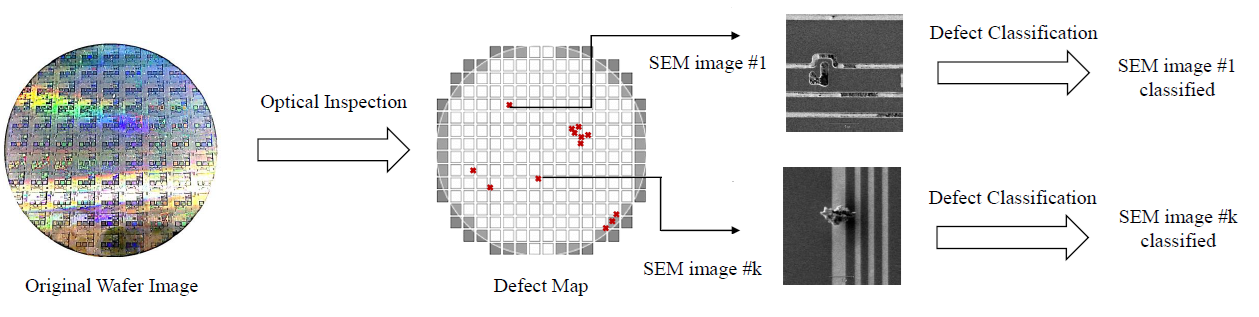}
\caption{Defect classification procedure for the semiconductor manufacturing process at hand: (i) an optical inspection tool locates defects (marked in red) by comparing multiple neighboring areas on a wafer and looking for anomalies; (ii) high-resolution SEM images are taken at each defect location. (iii) defects are classified: differences in the image background structure are visible, typically because images are taken in different positions on the wafer, or more generally, different wafers belong to different technologies, hence showing variations in their surface structure.}
\label{fig:pipeline}       
\end{figure*}

\begin{figure}[t]
  \centering
    \includegraphics[width=0.8\linewidth]{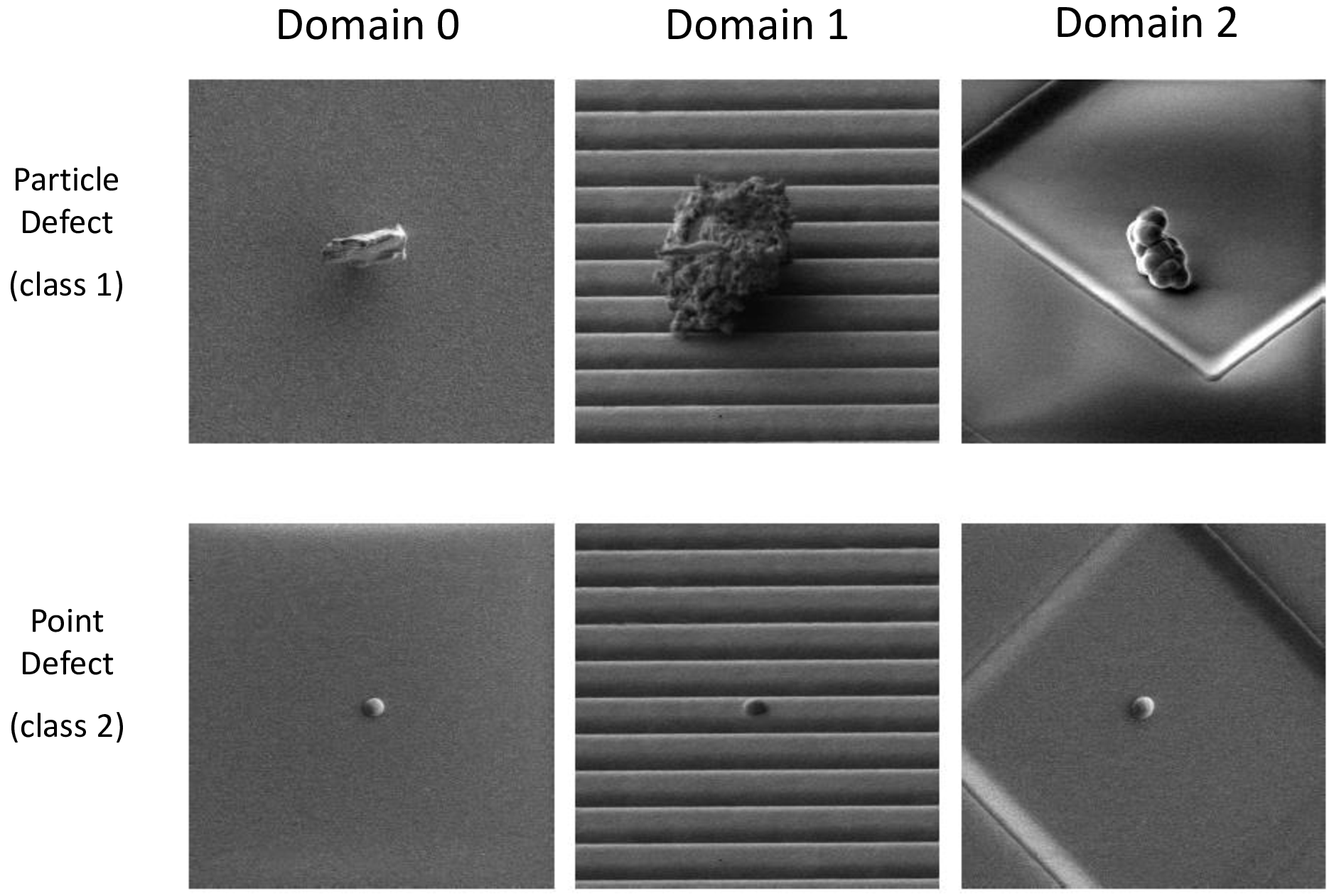}
    \caption{In this study, we address the image classification problem in the context of the semiconductor sector considering two classes: particle and point defects. Each column represents a distinct domain. Our objective is to apply Domain Adaptation techniques, using one domain as the source (labeled data) and another as the target (unlabeled data). The objective is to achieve adaptation to the new domain without the need for expensive label collection phases.}
    \label{fig:images_examples}
\end{figure}

Metrology is a set of quality assurance steps carried out in production to control and calibrate machines, process settings, process outcomes, and wafer properties. 
Within this framework, Defect Detection and Classification is a fundamental part of metrology to ensure high product quality and production performance. 
In this context, the use of Scanning Electron Microscope (SEM) images of the wafer can be used (see Figure \ref{fig:pipeline} and \ref{fig:images_examples}).
The collected data is examined and divided into different defect types before corrective actions or the declaration of ’defective’ products are triggered (the entire inspection procedure is depicted in Figure \ref{fig:pipeline}).

\add{
Industry 4.0, which is enabled by technologies such as Deep Learning, enabled automatic defect classification which has become cost-effective and widely adopted \cite{maggipinto2022deep}.
However, a series of challenges hinder the scalability of these models, such as the differences in data distributions combined with the abundance of data not labeled \cite{dalle2021formula}.
}

\add{
Domain Adaptation (DA) is extremely fitted for Industry 5.0 applications which focus on human-centric approaches, sustainability, and resilience.
Indeed, DA focuses on using the knowledge learned on a (source) domain to adapt and perform effectively on a different, but related (target) domain.
DA minimizes the need for extensive manual re-labeling or re-training of models, allowing human experts to focus on high-value tasks and improving robustness and scalability.
}

Specifically, in the context of the semiconductor field, to enable fab-wide usage of highly automated classification models, deployed methods need to be transferable and applicable when confronted with new optical properties and different image structures generated from technological diversity as well as reoccurring production steps in different stages of the manufacturing process.

\add{
Therefore, in this work, we propose to study the effectiveness of DA techniques in the context of the semiconductor domain.
In particular, we focus on state-of-the-art models able to perform in both the semi-supervised domain adaptation (SSDA) and the unsupervised domain adaptation (UDA) settings.
}
We focus on the \emph{AdaMatch} \cite{adamatch} approach for the pseudo-labeling group and \emph{DBACS} \cite{gentner2023} as a generative and adversarial model. 
While both have been proven to be of extreme applicability, they have never been applied so far in the context of image classification problems in semiconductor manufacturing.

In particular, in this work, we consider an evolution of the DANN approach \cite{dann} called \emph{DANN-based Alignment with Cyclic Supervision} (DBACS)\cite{gentner2023}.
This method was only considered for time-series data and it is studied for the first time in the image domain.
While this is a CycleGAN-inspired model, we enhance and adapt the original version for the computer vision by adding the following modifications: cyclic adaptation, unpaired sample mapping and feature matching loss, identity loss and multi-scale structure similarity loss.
\add{These modifications improved the final performance.}

The main contributions of this work are summarized as follows:
\begin{itemize}
\item We test and study for the first time the use of DA techniques to solve the defect classification problem in the semiconductor manufacturing field for the SEM images.
\item \manu{
We introduce the use of the DBACS approach in the Computer Vision domain by adapting its architecture and by considering additional loss terms.}
\item The effectiveness of our approach is validated on real-world Electron Microscope images considering the unsupervised and semi-supervised settings, proving also the usability and usefulness of DA techniques for the SEM images.
\end{itemize}

Furthermore, we have made all the code used in our study, including the DBCAS implementation, publicly accessible \footnote{ \url{https://bitbucket.org/papers_vad_group/dbacs/src/main/}}. This allows researchers and practitioners to  utilize our implementation for further advancements.

The rest of the paper is organized as follows: 
 In Section \ref{sec:literature}, we review the literature concerning the semiconductor sector and domain adaptation field. 
Section \ref{sec:our_approach} defines the problem formally and presents our approach.
Then Section \ref{section:proposed_approaches} describes the methods used as comparison.
 Section \ref{section:experimental_setting} provides detailed information about the data used, the experimental setup, including the considered models, and the specific hyperparameters employed for each DA technique. 
 Section \ref{section:results} shows and discusses the obtained results. 
 Finally, Section \ref{sec:conclusions} summarizes our findings and future research directions are presented,

\section{Related Work}\label{sec:literature}
We delineate the relevant literature as follows.
In Section \ref{sec:lit_image}, an overview of previous works to solve the image classification problem in semiconductor manufacturing is provided.
Then we discuss Domain Adaptation with a focus on the two main families, Pseudo-labeling in Section \ref{sec:lit_pseudo} and Generative Adversarial Models in Section \ref{sec:lit_domain}

	\begin{table}[h!]
		\begin{center}
			\begin{tabular}{c|c|c|c}
				Task & Labeled & Unlabeled & Distributions \\
				\hline \hline
				SSL & source & target & source $=$ target\\ \hline
				UDA & source & target & source $\ne$ target\\\hline
				SSDA & source + target & target & source $\ne$ target
			\end{tabular}
		\end{center}
		\caption{Comparison of different learning paradigm: Self-Supervised Learning (SSL), Unsupervised Domain Adaptation (UDA), Semi-Supervised Domain Adaptation (SSDA).}
			\label{tab:learning_paradigm}
	\end{table}
    
\subsection{Image Classification in Semiconductor Manufacturing}\label{sec:lit_image}
Image classification in semiconductor manufacturing is mainly applied for two different tasks: \emph{image defect classification} and  \emph{wafer-level/wafer map defect pattern classification}.
In image defect classification, images are taken at relevant spots in the wafer and then classified into apriori known defects (see SEM image in Figure \ref{fig:pipeline}).
In wafer-level/wafer map defect pattern classification, it is considered a map of the wafer where pre-identified anomalies/potential defects have been already identified and the task is to classify such maps w.r.t. known defect patterns (see defect map in Figure \ref{fig:pipeline}).
Some research focusing on wafer-level/wafer map defect pattern classification was published in recent years, eventually also driven by the wafer map dataset WM-811k made publicly available by \cite{wu2014wafer}.

Deep Learning exhibits considerable effectiveness in numerous Industry 4.0 and 5.0 applications, addressing a wide range of problems. For instance, in predictive maintenance, Deep Learning has become a preferred choice due to its high performance and automatic feature extraction capabilities \cite{lorenti2023predictive}.
Similarly, in image defect classification, Deep Learning approaches help to drastically reduce the amount of time spent by human operators in manually tagging images \cite{defect_cnn_1}. 
It also shows that CNN-based transfer learning methods can classify microscopic defect images with high accuracy. 
In \cite{schlosser2019novel} authors proposed a stacked hybrid CNN that exploits an attention mechanism. Also, in this case, CNNs and Deep Learning approaches are the typical choices \cite{defect_cnn_2, o2020deep} for performing the classification task. \cite{defect_cnn_2} additionally demonstrates that a single CNN model can extract effective features for defect classification without using additional feature extraction algorithms. 
Two publications exploiting pseudo-labeling are \cite{Liu2021DefectCO} and \cite{maps_semi}. 
However, only a few works apply really deep architectures in this context.
In this work, we consider the application of well-known DL models such as MobileNet, ResNet50, and Resnet101.

\subsection{Domain Adaptation}
Typically in DL, higher accuracy can be achieved when more data but also unlabeled data is available and used. When the unlabeled data is assumed to come from the same distribution as the labeled data we talk about Self-Supervised Learning (SSL). When, instead, it is assumed that the unlabeled data comes from a different distribution then we are in the UDA setting, where the notion of \textit{source} and \textit{target} \textit{domains} is used. In real-world applications, the setting called SSDA usually occurs, where the source and the target distributions differ and there is a portion, usually small, of target domain data that is labeled. 
A taxonomy of all the settings is reported in Table \ref{tab:learning_paradigm}.

\subsubsection{Pseudo-labeling}\label{sec:lit_pseudo}
In general,  Pseudo-labeling (PL) is a well-known and often employed technique for semi-supervised and self-supervised learning.
Pseudo-labeling \cite{pseudo-label} is first applied to SSL (learning paradigm, see Table \ref{tab:learning_paradigm}) and consists in producing artificial labels for unlabelled images and in training the model to predict the artificial labels when fed unlabeled images as input. It uses the model class prediction as a label to train against if it is above a certain probability threshold. 
Several methods have been proposed in the literature that exploit this concept.
\\
For example, \emph{AugMix} improves model robustness and uncertainty estimation by combining data modification in more detailed image augmentation with Jensen-Shannon divergence as consistency loss. 
FixMatch \cite{fixmatch} is a successful attempt to address and simplify SSL. It combines various semi-supervised learning techniques from  \cite{mean-teachers} and \cite{Hendrycks2020AugMixAS}. Similarly to ReMixMatch \cite{remixmatch}, it also leverages CTAugment and especially Cutout \cite{cutout} to produce heavily distorted, \emph{strongly augmented}, versions of an image and compare them to \emph{weakly augmented} ones.
AdaMatch \cite{adamatch} extends FixMatch by also taking into consideration the domain shift between the labeled and the unlabeled data, making it a suitable model for UDA (see Table \ref{tab:learning_paradigm}). Since this is the current state-of-the-art for semi-supervised and unsupervised domain adaptation for vision classification tasks, it is selected as one of our applied models. \manu{Indeed, since the considered method (DBACS) can perform SSDA and UDA, this is the best state-of-the-art model for this category to compare with}. A detailed description of AdaMatch is given in Section \ref{section:proposed_approaches}.
\manu{
AdaEmbed, \cite{adaembed} tackles Semi-Supervised Domain Adaptation (SSDA) and improves performance by learning a shared embedding space and generating accurate pseudo-labels. 
}
\subsubsection{Generative and Adversarial Models}\label{sec:lit_domain} 

When considering image-to-image (I2I) translation, pix2pix \cite{pix2pix} or conditional GAN (cGAN) \cite{cgan}, the Generator component of this GAN-based \cite{gan} models is an autoencoder-like network that maps an input image from a known source distribution to another target distribution. Such image translations are used for changing properties of single objects or style of images when supervised learning is done or \textit{paired} images from source and target are available.
In the case of unpaired I2I no such association between the input/source and output/target images is available. CycleGAN \cite{cyclegan} tackles this problem by introducing a second generator-discriminator couple and trying to learn a bijective mapping by translating the output of the original generator back to the original input space. A \textit{cycle-consistency} loss is introduced to compare an image from source or target domains with the same image after forward and backward translation.
\manu{The concepts of I2I tasks, cGANs, and CycleGANs can be used in the Domain Adaptation field, where source and target domains are different and it is possible to transform the samples from one domain to another. This allows a model, trained on the source domain samples, to perform well on the target domain samples thanks to this transformation}.

Based on domain adaptation theory, Domain Adversarial Neural Networks (DANN) \cite{dann} \manu{are introduced}. They target the model generalization issue when facing data differences originally within training and test data.
In their settings, an adversarial training approach similar to GANs is introduced.  
The DANN approach used to tackle the UDA problem for image classification focuses on the alignment between the source and target domains of the latent features extracted by the convolutional part of a deep neural network classifier, called the feature extractor. 
This alignment can be obtained by measuring, and thus improving, the discrepancy between the latent features distributions of the source and target domains. Different probability distance measures are adopted: \cite{tzeng2015simultaneous} uses the maximum mean discrepancy, \cite{sun2016deep} uses correlation alignment between domains, \cite{saito2018maximum} measures the discrepancy by adopting multiple task classifiers.
AD-Aligning \cite{ad-aligning} is another adversarial approach that emulates human-like generalization by combining adversarial training with source-target domain alignment. By pretraining with Coral loss and standard loss, AD-Aligning aligns target domain statistics with those of the pretrained encoder, preserving robustness while accommodating domain shifts. AVATAR \cite{avatar} instead introduces a domain-adaptive approach combining adversarial learning, self-supervised strategies, and sample selection for robust performance, especially in unsupervised domain adaptation scenarios.
Dann-based Alignment Model (DBAM) \cite{dbam} is an adversarial domain adaptation model inspired by DANN and semiconductor manufacturing modelling problems. Originally it is applied to time series data and used to solve a Virtual Metrology/Soft Sensing task, to improve the manufacturing process monitoring and its related operations by allowing interpretability and comparability of input and aligned output. Since it aligns the original input spaces, it is similar to I2I-based GANs with an autoencoder-like generator and the aligned features can still be compared and are physically interpretable. 

In this paper, an extended version called DANN-based Alignment with Cyclic Supervision (DBACS) \cite{gentner2023} introduced for heterogeneous domain adaptation on time series data is enhanced to be used for an unpaired image-to-image translation task. 
The model tackles the domain adaptation problem by transforming the source domain images to the target domain images. This allows a classifier trained on the source domain to correctly classify defects on the target domain since the input images are transformed to the source domain.
To the best of our knowledge, this is the first application of a domain adaptation method on defect classification in semiconductor manufacturing. A detailed description of the method can be found in Section \ref{section:proposed_approaches}

\section{Other approaches}
\manu{
Recent advancements in Unsupervised Domain Adaptation have focused on avoiding the need for source data on training, with methods such as ProxyMix \cite{proxymix}, which integrate a mixup regularization to enhance domain adaptation, improving model robustness against noisy pseudo labels and domain discrepancies. \cite{nacl} improves UDA by addressing the issues of inconsistent learning objectives and noisy pseudo labels. It enhances domain adaptation by simultaneously learning instance discrimination and semantic alignment across domains while removing false positive and negative pairs through topology-based selection. 
Instead, in the field of Semi-Supervised domain adaptation, \cite{inter-domain-mixup} introduces a cross-domain feature alignment strategy that incorporates label information into model adaptation. This approach enables better cross-domain feature alignment and reduces label mismatch. Furthermore, Neighborhood Expansion leverages high-confidence pseudo-labeled samples from the target domain, enriching the label diversity and consequently boosting the performance of the adaptation model.
Recently, the fields of Continual Learning and Domain Adaptation are overlapping, and \cite{cl-da} tackles domain adaptation with limited labeled target data by using a Conditional Variational Auto-Encoder (CVAE) and soft domain attention.
}

\section{Our Approach}
\label{sec:our_approach}

In the following part, a mathematical formalization of the addressed problem is given in Section \ref{subsec:problem_formalization}.
In Section \ref{subsec:generative_models}, we discuss foundational approaches, including the DANN-based Alignment Model (DBAM) and its extension, the DBACS method, highlighting their principles and architectures. Finally, in Section \ref{subsec:our_approach}, we detail our contributions to DBACS by integrating novel modifications, such as cyclic adaptation, unpaired sample mapping, and additional loss functions.

\subsection{Problem Formalization}
\label{subsec:problem_formalization}
In this work, the focus is to learn a model in a classification setting and to scale it from one labeled source domain to a target domain with a different data distribution where the data are partially labeled (SSDA) or completely unlabeled (UDA).
Let's define $X$ as the input space, $Y$ as the output space, and a \emph{domain} as a distribution over $X \times Y$. A learning algorithm is now provided with a data set $S$ drawn i.i.d. from a domain $D_S$ with $X_S \times Y_S$, $X_S \subset X$, $Y_S \subset Y$. For better differentiation, $D_S$ is called \emph{source domain}. In the SSL setting, we distinguish between labeled and unlabeled data and define $S:=SL \cup SU$ where $SU$ stands for the unlabeled sample subset and $SL$ for the labeled one. Without loss of generality for UDA and SSL, $SU = \emptyset$ since the source domain is assumed to be labeled. Hence
\begin{equation}
    S=\{X_{S},Y_{S}\}=\{X_{SL},Y_{SL}\}=\{x^i_S,y^i_S\}^n_{i=1} \sim D_S,
\end{equation}
with $n$ being the number of drawn samples (all labeled) and therefore $X_S=X_{SL} \subset X$, $Y_S=Y_{SL} \subset Y$. We assume to have two data sets. Let $T=TL \cup TU$ be the second data set $T$ drawn i.i.d. from a domain  $D_T$ called \emph{target domain} with a distribution over $X_T \times Y_T$, $X_T \subset X$, $Y_T \subset Y$, and consisting of unlabeled $TU$ and/or labeled $TL$ samples.
\begin{align}
    TL &=\{X_{TL},Y_{TL}\}=\{x^j_T,y^j_T\}^{m-l}_{j=1} \sim D_T; \\
    TU &=\{X_{TU}\}=\{x^j_T\}^{m}_{j=m-l+1} \sim D^X_T;
\end{align}
with $m$ being the number of drawn target samples and $l$ the number of labeled samples, therefore $X_{TL} \subset X_T \subset X$, $Y_{TL} \subset Y_T \subset Y$ and $X_{TU} \subset X_T \subset X$.
Let $D^{X}_S$, $D^{Y}_S$ and $D^{X}_T$, $D^{Y}_T$ be the marginal distributions of $D_S$ and $D_T$ over $X_S$, $Y_S$ and $X_T$, $Y_T$ respectively.

\subsection{DBAM and DBACS methods}
\label{subsec:generative_models}
\emph{Domain Adversarial Neural Networks} (DANN) \cite{dann} measure the domain discrepancy using a domain discriminator network combined with suitable probability distance metric and by employing adversarial training. 
DANN-based Alignment Model (\textbf{DBAM}) is based on the DANN approach and it is presented in \cite{dbam} as a GAN model. The idea is to keep the necessary information to allow interpretability and comparison of source and target while maintaining the high accuracy of a dedicated model trained on the source domain. It is originally applied on time series and stationary data: DBAM consists of 3 parts: a classifier/predictor $f_{DB}$ that solves the task, a generator -  also called aligner $F$ - that maps target domain to source domain to enable classifier usage for mapped target data, discriminator (called discriminator A later) that distinguish between source data and aligned target data. DBAM uses I2I-like translation meaning it works with the original feature spaces for both source and target. In more detail, DBAM is mapping source input data into the original target feature space, via an autoencoder-like generator network. Adversarial training is applied using the discriminator. 

\begin{figure*}
        \centering
	\includegraphics[width=0.9\linewidth]{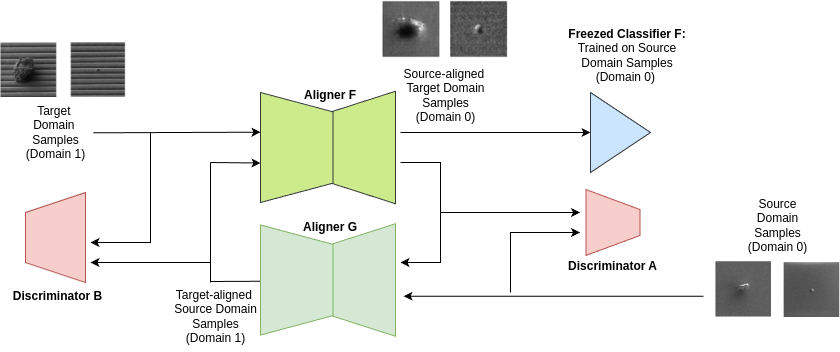}
	\caption{Scheme of our approach based on DBACS and enhanced with the following modifications: cyclic adaptation, unpaired sample mapping, identity loss, feature matching, and multi-scale structure similarity.
    The classifier is trained on source domain data. The first generator (aligner $F$) maps the target domain to the source domain to enable target data classification. Aligner $F$ combined with aligner $G$ are set up to enable bijective mapping within each domain. Discriminators are used for adversarial training. The discriminators and aligners compose the CycleGAN. The classifier, trained on the source domain, is frozen during the DBACS training. In this image example, Domain 0 is the Source Domain and Domain 1 is the Target Domain.} \label{fig:dbam_arch}
\end{figure*}

CycleGAN \cite{cyclegan}, a model designed for unpaired image-to-image translation by introducing cycle-consistency loss to map between two domains, is adopted and enhanced in this work.
In particular, we exploit an extended version of DBAM called DANN-based Alignment with Cyclic Supervision (\textbf{DBACS}) \cite{gentner2023}. Since the task asks for unpaired mapping, a second generator (called aligner $G$) is used that translates the source to the target domain and the output of aligner $F$ back to the target domain. Adversarial training in this direction is enabled by adding a second domain discriminator (called discriminator $B$). By combining both aligners it is possible to introduce \emph{cycle-consistency} by comparing source samples with its cycled sample and target samples with their cycled samples. A visualization of DBACS is presented in Figure \ref{fig:dbam_arch}.

\subsection{Our Approach}
\label{subsec:our_approach}
We extend the DBACS approach by using it for the computer vision domain.
In particular, we introduced the following series of modifications: cyclic adaptation, unpaired sample mapping, identity loss, feature matching, and multi-scale structure similarity.
We define the different parts of the final loss function $\mathcal{L}_{final}$ used for training. We apply the same training routine as for DBAM \cite{dbam} and DBACS \cite{gentner2023}, by using adversarial training, i.e. alternatively train discriminator and aligner parts with contradictory goals and a fixed ratio $r_{adv}$. The discriminator’s objective is to maximize the distance between the distributions of the source and the aligned target. The aligner’s objective is to minimize the distance between the source and aligned target distribution and to minimize the prediction error of the classifier model on the aligned target data. Let $f_{cc}: X \rightarrow Y$ be the on source data trained classification model with softmax output. 
Let $F: X_S \to X_T$ define a statistical model from source domain to target domain, $G: X_T \to X_S$ a statistical model from target to source domain, $D_A: X_S \to \{0,1\}$, $D_B: X_T \to \{0,1\}$ statistical model classifying source versus aligned target and target versus aligned source.
\subsubsection{DBACS - Classifier Loss} Only applicable in case of partly labeled target data when semi-supervised learning is possible. The classifier part of $\mathcal{L}_{final}$ is defined as:
\begin{align}
    \mathcal{L}_{cc}&=\mathcal{L}(f_{cc},Y_TL)\nonumber \\
    &= \sum_{\left(x_{TL},y_{TL}\right) \in T}\left(f_{cc}(F(x_{TL})),y{_{TL}}\right)
\end{align}
where $ \mathcal{L}$ is the categorical cross entropy loss. When used it supports task-specific alignment of target data.
\subsubsection{DBACS - Adversarial Loss} The adversarial loss is applied to the output of the Discriminator. During Generator training, it is minimized, during Discriminator training it is maximized (or its negative value minimized). First, we define it for the target to source alignment where F tries to map the target domain to the source domain, then for source to target alignment with $G$:
\begin{align}
    \mathcal{L}_{adv_{S}}\left(F, D_{A}, X_S, X_T\right) &= \log \left(1 - D_{A}(F(x_T))\right) + \nonumber \\ &+ \log \left(D_{A}(x_S)\right)  \\
    \mathcal{L}_{adv_{T}}\left(G, D_{B}, X_S, X_T\right) &= \log \left(1 - D_{B}(G(x_S))\right) + \nonumber \\ &+\log \left(D_{B}(x_T)\right) 
\end{align}
This is the original loss function used in GANs \cite{gan} and in DANN \cite{dann} for classification.
\subsubsection{DBACS - Cycle Consistency Loss} The cycle-consistency loss minimizes two differences: the one between source data $x_S \in X_S$ and their cycled version $F(G(x_S))$ and the one between target samples $x_T \in X_T$ and their cycled version $G(F(x_T))$:
\begin{equation} 
    \mathcal{L}_{cyc}(G, F) = \mathcal{L}_1(F(G(X_S)),X_S) + \mathcal{L}_1(G(F(X_T)),X_T)
\end{equation}
with $\mathcal{L}_1$ is chosen as L1-norm (see \cite{cyclegan}).
\subsubsection{DBACS - Identity Loss} Also used in \cite{cyclegan} in case of distribution overlap between source and target domain. Minimizes the differences between $x_T$ and $G(x_T)$ and between $x_S$ and $F(x_S)$ when fed with $x_T \in X_S$ and $x_S \in X_T$).
\begin{equation}
    \mathcal{L}_{id}(G, F) = \mathcal{L}_1(F(x_S),x_s) + \mathcal{L}_1(G(x_T),x_T).
\end{equation}
\subsubsection{DBACS - Additional Loss} Two additional losses inspired by \cite{ganimorph} are added. In case the mapping between the two domains is not bijective, the cyclic loss should aim to preserve the most important information by better preserving features visible to humans rather than noisy, high-frequency information. Hence \emph{Multi Scale Structure Similarity loss} (MS-SSIM) is added in addition to the cyclic loss.
\begin{align}
    \mathcal{L}_{ssim}(G, F) = & (1 -\mathcal{L}_{ssim}(F(G(X_S)),X_S)) \nonumber \\
    &+ (1 - \mathcal{L}_{ssim}(G(F(X_T)) ,X_T)).
\end{align}
\emph{Feature Matching loss} is introduced in order to encourage aligned and original samples to produce similar activation at each layer of the discriminator instead of just output layer.
\begin{align}
    \mathcal{L}_{fm_{A}}(F, D_{A}) &= \frac{1}{l - 1} \sum_{l=1}^{n - 1} \left\|a_{l}(X_S) - a_{l}(F(X_T))\right\|^{2}_{2} \\
    \mathcal{L}_{fm_{B}}(G, D_{B}) & = \frac{1}{l - 1} \sum_{l=1}^{n - 1} \left\|a_{l}(X_T) - a_{l}(G(X_S))\right\|^{2}_{2}
\end{align}
where $a_{l} \in D_{A}$ represents the raw activation of the $l^{th}$ layer of the discriminator $D_{A}$, and $l$ is the total number of discriminator's layers including output.

The final loss $\mathcal{L}_{final}$ is defined as:
\begin{align}
    \mathcal{L}_{final} = \lambda_{cc}\mathcal{L}_{cc} + \lambda_{adv}\mathcal{L}_{adv} &+ \lambda_{cyc}(\mathcal{L}_{cyc} + \mathcal{L}_{ssim})\nonumber  \\
    &+ \lambda_{id}\mathcal{L}_{id} + \lambda_{fm}\mathcal{L}_{fm}
\end{align}
with \begin{align*}\mathcal{L}_{adv} =\mathcal{L}_{adv_{S}} + \mathcal{L}_{adv_{T}}\\ 
\mathcal{L}_{fm} = \mathcal{L}_{fm_{A}} + \mathcal{L}_{fm_{B}}
\end{align*}
and where  $\lambda_{(\cdot)}$ represents the weight assigned to each corresponding loss term. We also follow some general architecture recommendations: both aligners have a U-Net \cite{unet} like architecture taken from \cite{pix2pix} but without cropping so that the skip connection could be applied. For both discriminators the architecture is taken from \cite{ganimorph}, which improves CycleGAN's shape deformation by introducing \emph{dilated convolutions}, often used in semantic segmentation: this allows a bigger receptive field with respect to the input image at the same cost of a convolution but with a much smaller filter.

\begin{algorithm}
	\caption{DBACS for defect classification} 
	\begin{algorithmic}[1]
 \label{alg:dbacs}
 \State Prepare source trained classifier $f_{cc}$ with frozen weights; two discriminators $D_A$,$D_B$; two aligners $F$, $G$; assign loss weights, learning rate, optimizer
		\For {$iteration=1,2,\ldots epochs $}
			\For {$steps=1,2,\ldots,ratio$}
				\State unfreeze $D_A$,$D_B$ weights, $F$, $G$ weights
				\State train $D_A$,$D_B$ by maximizing final loss $\mathcal{L}_{final}$ (minimizing $-\mathcal{L}_{final}$)
			\EndFor
			\State freeze $D_A$,$D_B$ weights, unfreeze $F$, $G$ weights
			\State train $F$, $G$ by minimizing final loss $\mathcal{L}_{final}$ , for SSDA including $\mathcal{L}_{cc}$
		\EndFor
	\end{algorithmic} 
\end{algorithm}

\section{Compared Approaches}
\label{section:proposed_approaches}
We discuss the first group of DA techniques called Pseudo Labeling in Section \ref{subsec:PseudoLabeling}.
Then we dedicated a detailed description of the state-of-the-art approach, AdaMatch in Section \ref{subsec:adamatch}.

\subsection{Pseudo-Labeling}
\label{subsec:PseudoLabeling}
\textbf{Pseudo-labeling} is a semi-supervised learning method: a neural network is trained on labeled source data by minimizing a supervised classification loss function. Then unlabeled data is used \add{to improve generalization and accuracy on the target domain}, hence further improving accuracy for source and, especially, for unlabeled target data. We assume softmax output unit, categorical cross entropy as a loss function $\mathcal{L}=\mathcal{L}_{CE}$. The goal is to build a statistical model $f_{ada}: X \rightarrow Y$ so that the classification error is minimized for $D_S$. For labeled source data, we have the following optimization problem:
\begin{align}
    f_{ada}(\cdot) &= \min_{\tilde{f}}\mathcal{L}(\tilde{f}(X_S),Y_S) \nonumber \\
    &=\min_{\tilde{f}} \sum_{\left(x^i_S,y^i_S\right) \in S} \mathcal{L}\left(\tilde{f}(x^i_S),y^i_{S}\right).
\end{align}
where $(x^i_S,y^i_S)$ is any paired source (input, output) sample drawn i.i.d from $D_S$.
$C$ represents the number of class labels, $y^k$ is the categorical encoding of the label, $f^k$ is the network output meaning the extended class probability for the $k$-th class label.

Pseudo-labels $\hat{y}_T$ are classes assigned to unlabeled target samples by using a classifier trained on $S=SL$ and selecting the class with maximum predicted probability:
\begin{equation}
    \hat{y}^j_T= \operatorname*{argmax} f_{ada}(x^j_T) 
\end{equation}
The process of generating pseudo-labels can be accomplished in a so called \textit{online} and an \textit{offline} manner:
\begin{itemize}
    \item \emph{Offline PL}: A classifier is trained with the available labeled data for a fixed amount of iterations. A probability threshold is defined. Then the classifier is used to generate pseudo-labels for unlabeled target data. Pseudo-labels with a softmax output higher than the threshold are selected, added to the labeled data, and used for the next training phase to further improve the accuracy of the classifier model, especially for target data samples.
    \item \emph{Online PL}: The thresholding and pseudo-labeling are being performed in a temporary manner on each mini-batch during the classifier training. A loss weight based on the number of training steps is introduced to account for the lower classification accuracy, especially at the beginning of the training. Increasing the weight during is recommended (see \cite{fixmatch}). The loss function $\mathcal{L}$ is defined as:
\begin{equation}
    \mathcal{L}=\mathcal{L}(f_{ada}(X_S),Y_S) + \alpha(t) \mathcal{L}(f_{ada}(X_T),\hat{Y}_T),
\end{equation}
where $\alpha(t)$ is the weighting function schedule based on total number of steps $T$.
\end{itemize}

\subsection{AdaMatch}
\label{subsec:adamatch}
\textbf{AdaMatch} \cite{adamatch} exploits online pseudo-labeling and enriches it with consistency regularization, addressing the distribution shift between source and target domains present in the \emph{batch norm statistics}, flexible pseudo-label confidence threshold and modified version of distribution alignment. 
Let define $f: X \to Y=R^C$ the classifier model as before, where f is composed as $f(x)=g((h(x))$.
Specifically $h:X \to Z=R^C $  $f_1: X\to Z:=R^C$, where $Z$ represents the logits in output for each of the C classes. Instead, $g$ is defined as the softmax operator, which allows f to map the input in the probability space.

Then, AdaMatch follows the following pipeline.
\subsubsection{AdaMatch - Image Augmentation} Each minibatch $X_{SL}, X_{TU}$ are augmented both with a \emph{weak} and a \emph{strong} augmentation,  $X^{aug}_{SL}=\{X_{SL, w},X_{SL, s}\}$, $X^{aug}_{TU}=\{X_{TU, w},X_{TU, s}\}$ respectively. The logits of the augmented images are computed as:
\begin{align}
    \{Z'_{SL}, Z_{TU}\} & = h_{ada}(X^{aug}_{SL}, X^{aug}_{TU})\nonumber \\
    Z''_{SL} & = h_{ada}(X^{aug}_{SL}), 
\end{align}
where $Z'_{SL}$ and $Z''_{SL}$ could be different because of the batch normalization statistics since they are obtained from differently assembled batches (respectively with and without the target domain images).
\subsubsection{AdaMatch - Random Logit Interpolation} The logits $Z'_{SL}$ and $Z''_{SL}$ are interpolated:
\begin{equation} 
    Z_{SL} = \lambda Z'_{SL} + (1 - \lambda) Z''_{SL},
\end{equation}
where $\lambda$ is sampled from a uniform distribution with range $(0, 1)$ to weight each logit.
\subsubsection{AdaMatch - Distribution Alignment} The idea is to encourage the target unlabeled distribution of pseudo-labels to follow the distribution of the source labeled data assuming similar source and target label distributions. 
AdaMatch estimates the source label distribution from the output of the model on the labeled source data. Given the logits for weakly augmented source $Z_{SL,w}$ and target $Z_{TU,w}$ samples, the pseudo-labels are computed:
\begin{align*}
	\label{eq:dist_alignment}
	\hat{Y}_{SL,w} &= \operatorname{softmax}(Z_{SL,w})\nonumber \\
	\hat{Y}_{TU,w} &= \operatorname{softmax}(Z_{TU,w}).
\end{align*}
Distribution alignment is applied by multiplying the target unlabeled pseudo-labels by the ratio between the expected value of the weakly augmented source pseudo-labels $\mathbb{E}[\hat{Y}_{SL,w}]$ and the expected value of the weakly augmented target pseudo-labels $\mathbb{E}[\hat{Y}_{TU,w}]$:
\begin{equation}
    \tilde{Y}_{TU,w} = \operatorname{normalize}\left(\hat{Y}_{TU,w}\frac{\mathbb{E}[\hat{Y}_{SL,w}]}{\mathbb{E}[\hat{Y}_{TU,w}]}\right), 
\end{equation}
where $\operatorname{normalize}$ makes the distribution sum to 1 again.

\subsubsection{AdaMatch - Loss Function} The loss for the source labeled, both weakly and strongly augmented data is defined as:
\begin{align}
     \mathcal{L}_{source} = \frac{\tau}{n_{SL}} \sum_{i=1}^{n_{SL}} \mathcal{L}\left({Y}^{(i)}_{SL}, {Z}^{(i)}_{SL,w}\right) + 
     \\ + \frac{\tau}{n_{SL}} \sum_{i=1}^{n_{SL}} \mathcal{L}\left({Y}^{(i)}_{SL}, {Z}^{(i)}_{SL,s}\right)
     \nonumber 
     \end{align}
The loss for the unlabeled, masked target data is defined as:
\begin{align} 
    \mathcal{L}_{target} = \frac{\tau}{n_{TU}} \sum_{i=1}^{n_{TU}} \mathcal{L}\left(\operatorname{stop\_grad}(\tilde{Y}^{(i)}_{TU,w}), {Z}^{(i)}_{TU,s}\right) mask^{(i)}
\end{align}
where $\mathcal{L} = \mathcal{L}_{CE}$ is the categorical cross entropy loss and $\operatorname{stop\_grad}$ is a function that prevents gradient back-propagation on pseudo-labels, which is a standard practice in SSL that improves training. The final loss $\mathcal{L}_{final}$ is then:
\begin{equation} 
    \mathcal{L}_{final} = \mathcal{L}_{source} + \mu(t) \mathcal{L}_{target}
\end{equation}
where $\mu (t)$ is a warm-up function.

\subsubsection{AdaMatch - Augmentation}
In AdaMatch, each source and target image is augmented both in a weak and a strong way. Each weakly augmented image is randomly horizontally mirrored with a certain probability.
The strong augmentation is given by CTAugment \cite{remixmatch} applied on top of the same horizontal flip. CTAugment is based on AutoAugment \cite{autoaugment}, it randomly selects a number of transformations for each sample and learns the magnitudes of each individual transformation on-the-fly. 
In particular, given a collection of transformations (augmentations), each sample of a mini-batch is augmented with a pipeline consisting of two transformations that are randomly and uniformly sampled; Cutout \cite{cutout}, by a square as big as 1/4 of the image size by 1/4 of the image size, with an area of 1/16 of the total area, is always an additional part of the strong augmentation pipeline and applied on top of the two augmentations. 
The magnitude of each transformation is first chosen randomly and then updated according to how close the model's predictions are to the true labels. The set of augmentations is given by: autocontrast, brightness, color, contrast, cutout, equalize, invert, identity, posterize, rescale, rotate, sharpness, shear\_x, shear\_y, smooth, solarize, translate\_x, translate\_y.  
Specific modifications of the augmentation procedure with respect to the original paper are as follows.
In order to avoid the introduction of new artifacts that are not present in the original data domain, only 180 degree rotations are applied (especially important for horizontal lines in domain 1). In addition, the size of the cutouts is selected small enough to avoid complete cutouts of any whole defect.

\section{Experimental Setting}
\label{section:experimental_setting}

\begin{figure*}
        \centering
	\includegraphics[width=\linewidth]{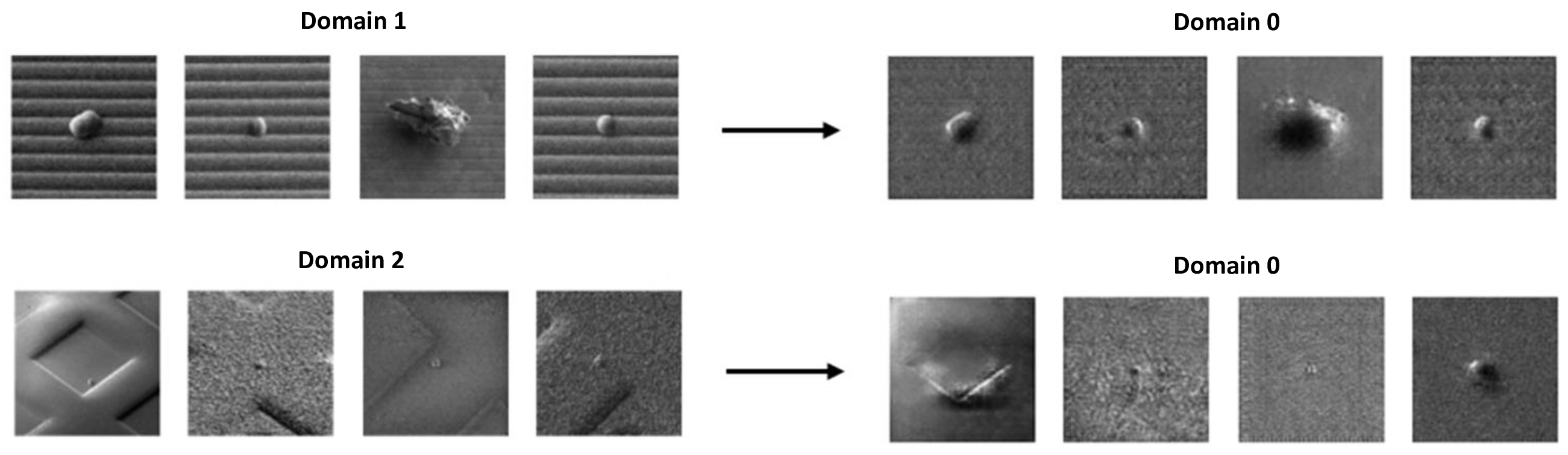}
	\caption{Examples of defect images aligned to Domain 0 with DBACS in the UDA setting. The left column shows the original images; the right column shows the aligned images with changed backgrounds. The first row shows domain 1, and the second row shows domain 2.}
	\label{fig:dbam_aligned_images}
\end{figure*}

\subsection{Dataset Description}
\label{section:dataset}
The data described in this section and used in the experiments have been provided by Infineon Technologies AG; the dataset is composed of images taken by SEMs. As introduced in Section \ref{sec:intro}, SEM is a type of electron microscope that produces images of a specimen by scanning its surface with a focused beam of electrons. The electrons, by interacting with atoms in the specimen, produce several informative signals about the surface topography and composition of the specimen. Secondary electrons emitted by the specimens' atoms excited by the electron beam can be detected by using an in-lens detector or an external detector. Depending on the type of detector, SEM images can therefore be divided into in-lens detector images and external detector images. In this work, both kinds of images are used.

The data includes different predefined \emph{product technologies} that are characterized by different backgrounds on which the defects lie. Only defects are selected that can occur on multiple product technologies due to overlapping production steps. Therefore, a subset of the classes is chosen, and occurring defects are merged into two final classes, which are called \emph{points} and \emph{particle}. A particle defect is characterized by some kind of foreign particle, like, for example, dust, that is present on the current surface layer of the wafer. Point defects are impurities of the layer material and can disturb physical properties like, for example, the electrical conductivity of the wafer. Three different product technologies presenting three different image backgrounds are selected based on the availability of training data. 
The three domains are defined based on the corresponding product technologies: domain 0 has a plain, unstructured background; \emph{domain 1} has horizontal lines in the background; \emph{domain 2} exhibits square-like shapes of different sizes and locations in the background.
More in detail, all images are grayscale of shape 128x128, where for Domain 0, there are 1706 instances in the first class and 1516 instances in the other.
Domain 1 has 639 and 503 instances in its respective classes.
In domain 2, there are 577 instances in one class and 1196 in the other.
It is also relevant to observe that the class distribution is balanced for domains 0 and 1 but imbalanced for domain 2.

\begin{table*}[th!]
    \begin{center}
        \caption[Baseline models accuracy]{\textbf{Baseline models accuracy.} The first column denotes the domain of the train data, and the remaining columns indicate the accuracy of the corresponding test set while changing the model. In \textbf{bold} the best value for each domain and model.}
        \label{tab:baselines_acc}
        \addtolength{\tabcolsep}{-3pt} 
        \small 
        \begin{tabular}{|c|*{9}{c|}} 
            \hline
            \textbf{} & \multicolumn{3}{c|}{\textbf{Test Domain 0}} & \multicolumn{3}{c|}{\textbf{Test Domain 1}} & \multicolumn{3}{c|}{\textbf{Test Domain 2}} \\
            \hline
            \textbf{Training Domain} & \textbf{MobileNet} & \textbf{Resnet50} & \textbf{Resnet101} & \textbf{MobileNet} & \textbf{Resnet50} & \textbf{Resnet101} & \textbf{MobileNet} & \textbf{Resnet50} & \textbf{Resnet101} \\
            \hline
            \textbf{0} & \textbf{0.976} & \textbf{0.921} & \textbf{0.981} & 0.707 & 0.680 & 0.810 & 0.662 & 0.657 & 0.669 \\
            \hline
            \textbf{1} & 0.672 & 0.733 & 0.699 & \textbf{0.973} & \textbf{0.858} & \textbf{0.982} & 0.512 & 0.574 & 0.583 \\
            \hline
            \textbf{2} & 0.841 & 0.629 & 0.876 & 0.828 & 0.677 & 0.799 & \textbf{0.959} & \textbf{0.811} & \textbf{0.971} \\
            \hline
            0 + 5\% 1 & 94,06 & 95,66 & 94,52 & 88,71 & 90,82 & 92,84 & 72,55 & 82,53 & 84,54 \\
            \hline
            0 + 5\% 2 & 93,12 & 94,44 & 94,8 & 83,19 & 75,43 & 84,11 & 87,96 & 86,27 & 87,3 \\
            \hline
        \end{tabular}
    \end{center}
\end{table*}

\subsection{UDA and SSDA Scenarios}
\label{subsec:uda_ssda}
We are going to simulate unsupervised domain adaptation (UDA) as well as semi-supervised domain adaptation (SSDA). To simulate an UDA scenario, a source domain is selected and assumed to be fully supervised; hence, all available labels will be considered during training. One of the other domains is selected as the target domain, and all available labels are ignored and not used for training. For example, for UDA from domain 0 to 1, the labels of the samples from domain 0 are considered, while those from domain 1 are ignored. In the SSDA scenario, a pre-defined percentage of samples (from both classes) from the defined target domain is labeled and available during the training phase. For evaluation of model accuracy, available labels in the corresponding test data sets are used.
\\
In the UDA and SSDA scenarios, we evaluate and compare the following DA techniques: Offline and Online PL, AdaMatch, and DBACS.
These results will also be compared against an \textit{Oracle} (upper bound) and a \textit{lower bound} (baseline). The Oracle represents the performance reported in Table \ref{tab:baselines_acc} when trained on source domain $i$ and tested on the target domain $i$. Essentially, the Oracle is the model trained with the labeled data of the studied domain.
The \textit{lower bound} for a target domain $j$ using the source domain $i$ is defined by the values in Table \ref{tab:baselines_acc}, where the model is trained on domain $i$ and evaluated on domain $j$.

\subsection{Models}
\label{subsec:models}
Several DA techniques are tested in this work, like Online and Offline PL, including also state-of-the-art AdaMatch and other techniques like DBACS.
In this work, we examine deeply the different approaches by comparing them while also changing the underlying classifier architecture.
Specifically, we test all the DA methods with the following three implemented architectures: MobileNet, Resnet50, Resnet101.
We performed this additional examination to evaluate the robustness of DA methods respect to architectural changes.
Moreover, it is relevant to understand if the use of bigger networks like Resnet101 or lighter architectures like MobileNet can influence the final performance \cite{sarraf2021comprehensive}.
All the models are implemented in Python by using Tensorflow\footnote{https://www.tensorflow.org}.
\\
For all architectures, the pre-trained model from ImageNet is evaluated considering only the feature extractor part.
Then a binary classifier head suitable for the task at hand is attached, made by one dense layer. 
Since the network was pretrained on colored images (hence with a 3-channel input) and considering that the defect images are grayscale; it is more convenient to work with one channel only, a Conv2D layer with 3 filters was added on top as first and new input layer.
\\
The newly created layers are first trained for 30 epochs with the learning rate set to 0.0001 and Adam used as optimizer. While the first layers of the feature extractor remain frozen, the remaining layers are trained for 20 more epochs, with a learning rate that is 1/10 of the initial learning rate. Both phases of the training employ $20\%$ the training data for validation purposes; early stopping with patience 5 is applied on the validation loss.

\subsection{Hyperparameter Setting for the DA methods}
\label{subsec:methods}
All methods, unless diversely specified, are trained with the same hyperparameters as  the classifier baselines.

\subsubsection{Offline and Online PL}
Both Offline and Online PL consider a confidence threshold $\tau = 0.9$ and $r=3$ as unlabeled to labeled sample ratio. The offline ST considers $N = 10$ iterations. 

\subsubsection{AdaMatch}
For the AdaMatch approach, our results showed that blind usage of the default hyperparameters leads to suboptimal performance.
Instead, tuning the data augmentations used and taking into account the effect of the different label distributions across domains can make it the most competitive approach for some source-target distributions.
AdaMatch uses, $\tau = 0.9$ and $r=3$ like the PL approaches.
Then it is used a weight decay of $0.001$, a learning rate of $0.0002$, and a batch size of $64$. 
Given the limited amount of data, we reduce the number of steps by a factor 8 with respect to the original paper.
In order to compensate for the reduced training, the weight of the unsupervised loss is also adapted to prevent too much weight to the unsupervised loss too early in the training.

\subsubsection{DBACS}
The adversarial training of the other model parts of the architecture (discriminators and aligners) is carried out with a training ratio $r_{adv}=2$ in favour of the discriminators. The final batch size is $64*r_{adv}=128$, the learning rate $lr=0.00005$, the optimizer is Adam, and the number of epochs is $300$. The weights assigned to the different loss terms are $\lambda_{ce}=1.0, \lambda_{adv}=0.5, \lambda_{cyc}=0.3, \lambda_{id}=0.2, \lambda_{fm}=0.0$ and follow the indications in \cite{ganimorph},

\FloatBarrier

\section{Results} 
\label{section:results}

The DA methods and baselines are evaluated for the UDA setting in Section \ref{subsec:uda}, while in Section \ref{subsec:ssda} a similar evaluation is performed for the SSDA setting.

\begin{table*}[th!]
\begin{center}
\caption[UDA models accuracy - source domain 0]{\textbf{UDA models accuracy - source domain 0.} In \textbf{bold} the best value for each domain and model.}
\label{tab:uda_acc_source_0}
  
\begin{tabular}{|c|ccc|ccc|}
\hline
Source $\rightarrow$ Target & \multicolumn{3}{c|}{0 $\rightarrow$ 1}                                                            & \multicolumn{3}{c|}{0 $\rightarrow$ 2}                                                             \\ \hline
Model                       & \multicolumn{1}{c|}{MobileNet}      & \multicolumn{1}{c|}{ResNet50}    & ResNet101     & \multicolumn{1}{c|}{MobileNet}   & \multicolumn{1}{c|}{ResNet50}       & ResNet101      \\ \hline
lower limit                 & \multicolumn{1}{c|}{70,7}           & \multicolumn{1}{c|}{68}          & 81            & \multicolumn{1}{c|}{66,2}        & \multicolumn{1}{c|}{65,7}           & 66,9           \\ \hline
DBACS                       & \multicolumn{1}{c|}{79,3}           & \multicolumn{1}{c|}{72,4}        & 80,2          & \multicolumn{1}{c|}{64,53}       & \multicolumn{1}{c|}{61}             & 65,64          \\ \hline
Offline PL                  & \multicolumn{1}{c|}{83}             & \multicolumn{1}{c|}{\textbf{86}} & 85            & \multicolumn{1}{c|}{55,32}       & \multicolumn{1}{c|}{52,78}          & 53,72          \\ \hline
Online PL                   & \multicolumn{1}{c|}{67,8}           & \multicolumn{1}{c|}{57}          & 69,5          & \multicolumn{1}{c|}{59,8}        & \multicolumn{1}{c|}{69,6}           & 69,8           \\ \hline
AdaMatch                    & \multicolumn{1}{c|}{\textbf{81,89}} & \multicolumn{1}{c|}{84,15}       & \textbf{85,3} & \multicolumn{1}{c|}{\textbf{79}} & \multicolumn{1}{c|}{\textbf{81,23}} & \textbf{84,32} \\ \hline
Oracle                      & \multicolumn{1}{c|}{97,3}           & \multicolumn{1}{c|}{85,8}        & 98,2          & \multicolumn{1}{c|}{95,9}        & \multicolumn{1}{c|}{81,1}           & 97,1           \\ \hline            
\end{tabular}
\end{center}
\end{table*}

\begin{table*}[th!]
\begin{center}
		\caption[UDA models accuracy - source domain 3]{\textbf{UDA models accuracy - source domain 1.} In \textbf{bold} the best value for each domain and model.}
		\label{tab:uda_acc_source_3}
\begin{tabular}{|c|ccc|ccc|}
\hline
\textbf{Source $\rightarrow$ Target} & \multicolumn{3}{c|}{\textbf{1 $\rightarrow$ 0}}                                                       & \multicolumn{3}{c|}{\textbf{1 $\rightarrow$ 2}}                                                \\ \hline
\textbf{Model}                       & \multicolumn{1}{c|}{\textbf{MobileNet}} & \multicolumn{1}{c|}{\textbf{ResNet50}} & \textbf{ResNet101} & \multicolumn{1}{c|}{\textbf{MobileNet}} & \multicolumn{1}{c|}{ResNet50}       & ResNet101      \\ \hline
\textbf{lower limit}                 & \multicolumn{1}{c|}{67,2}               & \multicolumn{1}{c|}{73,3}              & 69,9               & \multicolumn{1}{c|}{51,2}               & \multicolumn{1}{c|}{57,4}           & 58,3           \\ \hline
\textbf{DBACS}                       & \multicolumn{1}{c|}{67,84}              & \multicolumn{1}{c|}{69,53}             & 78,75              & \multicolumn{1}{c|}{55,18}              & \multicolumn{1}{c|}{63,24}          & 65,27          \\ \hline
\textbf{Offline PL}                  & \multicolumn{1}{c|}{\textbf{85,78}}     & \multicolumn{1}{c|}{\textbf{91,38}}    & \textbf{88,79}     & \multicolumn{1}{c|}{\textbf{82,64}}     & \multicolumn{1}{c|}{\textbf{85,64}} & \textbf{87,65} \\ \hline
\textbf{Online PL}                   & \multicolumn{1}{c|}{72,3}               & \multicolumn{1}{c|}{74,51}             & 76,01              & \multicolumn{1}{c|}{71,44}              & \multicolumn{1}{c|}{73,56}          & 72,04          \\ \hline
\textbf{AdaMatch}                    & \multicolumn{1}{c|}{56,65}              & \multicolumn{1}{c|}{60,78}             & 65,74              & \multicolumn{1}{c|}{52,86}              & \multicolumn{1}{c|}{54,33}          & 60,24          \\ \hline
\textbf{Oracle}                      & \multicolumn{1}{c|}{97,6}               & \multicolumn{1}{c|}{92,1}              & 98,1               & \multicolumn{1}{c|}{95,9}               & \multicolumn{1}{c|}{81,1}           & 97,1           \\ \hline
\end{tabular}
\end{center}
\end{table*}

\begin{table*}[!h]
	\begin{center}
		\caption
		{\textbf{SSDA ($5\%$ of labeled data in the target domain) models accuracy.} In \textbf{bold} the best value for each domain and model.}
		\label{tab:ssda_acc}
\begin{tabular}{|c|ccc|ccc|}
\hline
\textbf{Source $\rightarrow$ Target} & \multicolumn{3}{c|}{\textbf{0 $\rightarrow$ 1}}                                                       & \multicolumn{3}{c|}{\textbf{0 $\rightarrow$ 2}}                                                \\ \hline
\textbf{Model}                       & \multicolumn{1}{c|}{\textbf{MobileNet}} & \multicolumn{1}{c|}{\textbf{ResNet50}} & \textbf{ResNet101} & \multicolumn{1}{c|}{\textbf{MobileNet}} & \multicolumn{1}{c|}{ResNet50}       & ResNet101      \\ \hline
\textbf{lower limit}                 & \multicolumn{1}{c|}{88,71}              & \multicolumn{1}{c|}{90,82}             & \textbf{92,84}     & \multicolumn{1}{c|}{\textbf{87,96}}     & \multicolumn{1}{c|}{86,27}          & 87,3           \\ \hline
\textbf{DBACS}                       & \multicolumn{1}{c|}{81,67}              & \multicolumn{1}{c|}{82,74}             & 82,86              & \multicolumn{1}{c|}{70,03}              & \multicolumn{1}{c|}{77,87}          & 79,83          \\ \hline
\textbf{Offline PL}                  & \multicolumn{1}{c|}{\textbf{89,22}}     & \multicolumn{1}{c|}{\textbf{91,4}}     & 92                 & \multicolumn{1}{c|}{87,68}              & \multicolumn{1}{c|}{83,95}          & \textbf{89,09} \\ \hline
\textbf{Online PL}                   & \multicolumn{1}{c|}{87,4}               & \multicolumn{1}{c|}{88,35}             & 89,44              & \multicolumn{1}{c|}{82,45}              & \multicolumn{1}{c|}{\textbf{87,68}} & 88,78          \\ \hline
\textbf{AdaMatch}                    & \multicolumn{1}{c|}{88,1}               & \multicolumn{1}{c|}{89,11}             & 90,12              & \multicolumn{1}{c|}{85,25}              & \multicolumn{1}{c|}{86,75}          & 87,13          \\ \hline
\textbf{Oracle}                      & \multicolumn{1}{c|}{97,3}               & \multicolumn{1}{c|}{85,8}              & 98,2               & \multicolumn{1}{c|}{95,9}               & \multicolumn{1}{c|}{81,1}           & 97,1           \\ \hline
\end{tabular}
\end{center}
\end{table*}

\subsection{UDA scenario} 
\label{subsec:uda}
In the unsupervised setting, we have access to the label for the source domain but not for the target.
\\
As Oracle (optimal performance) is considered the performance when the model is trained on the labeled domain.
The results for each domain and for each model are reported in Table \ref{tab:baselines_acc}.
In general we can observe that the best values are achieved by the ResNet101 (largest model) followed by the MobileNet (lightest model) and interestingly ending with the ResNet50. 
Therefore, while the largest model like ResNet seems to be preferred, satisfactory results can be achieved with lighter architectures like MobileNet.
\\
When considering the \textit{lower limit} as baseline in terms of transfer learning, we consider the out-of-domain accuracy reported in Table \ref{tab:baselines_acc} and as described in \ref{subsec:uda_ssda}.
When considering these values, it is interesting to note that both ResNet101 and MobileNet seem to have a discrete ability of generalization.
\\
Based on the considered source domain is normal to expect different performance in terms of domain adaptation.
We report the results obtained by using as source the domain 0 in Table \ref{tab:uda_acc_source_0}, while in Table \ref{tab:uda_acc_source_3} we report the results for target domains 0 and 2 using as source the domain 1.
\\
The first general observation is that for the UDA setting, the DA techniques prove their usefulness in the semiconductor sector with SEM images by obtaining better performance than the lower limit (baseline).
This first observation confirms that the use of DA techniques can be beneficial in practice when we don't have access to labeled data of the target domain or when it is too costly to obtain the labels of the new domain.
\\
The second observation is that the best DA approach depends on the chosen source domain and the best DA approach depends on the specific use case.
In other words, the effectiveness of these methods can be highly dependent on the characteristics and distributions of the source data and all the tested approaches should be considered before choosing the best one.
\\
Another general observation concerns the application for the first time of the DBACS approach in the Computer Vision field (after all the modifications (as indicated in Section \ref{subsec:problem_formalization})) concerning the original approach to improve the results. 
More in detail, concerning the DBACS approach, in Figure \ref{fig:dbam_aligned_images}, examples of defect images from domains 1 and 2 using as the source domain 0.
Based on the results, we can note how the approach tries to change the data from the unlabeled domain into an already learned representation (domain 0).
DBACS demonstrates superior performance compared to the lower limit (baseline), highlighting its relevance to be used in future studies in the Domain Adaptation field.
Another advantage of the DBACS approach is that it remains unaffected by the source or target distribution, in contrast to the AdaMatch approach, which, in the case of relevant differences between the two distributions, could lead to low performance.
\manu{
However, it should be noted that DBACS is more computationally demanding compared to AdaMatch, especially given its complexity in terms of both architecture and optimization. 
}
\\
Examining the results from source domain 0 in Table \ref{tab:uda_acc_source_0}, it is evident that AdaMatch emerges as the top-performing approach. However, other methods such as DBACS and Offline PL also exhibit strong performance, notably surpassing the lower limit (baseline) by a considerable margin.
\\
In contrast, when observing the results from Table \ref{tab:uda_acc_source_3} concerning the source domain 1 we can see that most of the DA approaches struggle to achieve performance comparable to the Oracle (upper bound).
This includes the AdaMatch and DBACS methods. 
However, the best performer is offline PL, showing a significant improvement over the lower limit. 
For instance, in the case of $1 \rightarrow 0$ using ResNet 101, we achieve an accuracy of 88.79\% with Offline PL compared to 69.9\% with the lower limit.
This can be explained by the fact that more data are pseudo-labeled after each iteration. 
This means that the model is more confident about its predictions on unlabeled target domain data, as more target domain new data is available for training of the model.


\subsection{SSDA scenario}
\label{subsec:ssda}
In the SSDA scenario (a small part of the target domain data is available), the label availability improves the inter-domain performance of the DA models significantly (see Table \ref{tab:ssda_acc}).
This implies, in general, the importance of having even a small set of labeled data when possible.
However, at the same time, it hardly improves the performance concerning the "lower limit" baseline.
This raises concerns about the effectiveness of domain adaptation (DA) techniques in the semi-supervised domain adaptation (SSDA) scenario for the SEM images in the semiconductor manufacturing field, while in contrast, for the UDA scenario, we can observe a large benefit in the application of DA techniques.
Also, in this case, the best DA approach is Offline PL, though in some cases the lower limit has better results.

\section{Conclusion and Future Work}\label{sec:conclusions}
Training models able to generalize on out-of-distribution data can be a complex task. In fact, models trained on data coming from one domain struggle to generalize on other domains where labeled data is not accessible. 
In this work, Pseudo-labeling techniques and generative adversarial methods for the Domain Adaptation field are applied for the first time in the semiconductor manufacturing field for SEM images.
Different approaches are adopted and applied: the generative adversarial one represented by DBACS, which is meant to transform respective align input data from one domain to another, or the pseudo-label-based one, such as the AdaMatch model, is implemented and specified.
The results obtained show how the DA techniques are able to work effectively in the unsupervised setting (UDA), with large improvements over the baselines and results that are not far from the oracle's performance. 
Moreover, from this analysis, it is observed that the best DA approach depends on the chosen source domain and by the specific use case.
In other words, the effectiveness of these methods can be highly dependent on the characteristics and distributions of the source data and all the tested approaches should be considered before choosing the best one.
In addition, the initial tests conducted in the Computer Vision field using the DBACS approach appear promising, demonstrating the method's applicability and suggesting its potential for future studies in the Domain Adaptation field, especially in the case of distribution misalignments between the source and target distribution.
\\
As future research, we envision a focus on studying classification problems with more than two classes by including more defect classes, the impact of a larger amount of data, both labeled and unlabeled, and also on studying a multi-target domain adaptation scenario.

\FloatBarrier

\bibliographystyle{unsrtnat}  
\bibliography{Bibliography}

\end{document}